\documentclass{article}

     \usepackage[nonatbib, preprint]{neurips_2020}
\usepackage{environ}
\usepackage[utf8]{inputenc} 
\usepackage[T1]{fontenc}    
\usepackage{hyperref}       
\usepackage{url}            
\usepackage{booktabs}       
\usepackage{amsfonts}       
\usepackage{nicefrac}       
\usepackage{microtype}      
\usepackage{cite}
\usepackage{mathrsfs}
\usepackage{amsmath}
\usepackage{graphicx}
\usepackage{subfigure}
\usepackage[belowskip=-6pt]{caption}
\usepackage{graphicx}
\usepackage{amsmath}
\usepackage{amsthm}
\usepackage{amssymb}
\usepackage{amsfonts}
\usepackage{color}

\newcommand\T{\rule{0pt}{2.6ex}}       
\newcommand\B{\rule[-1.2ex]{0pt}{0pt}} 

\title{Deep Multimodal Transfer-Learned Regression in Data-Poor Domains}

\author{%
Levi D. McClenny\\ 
Dept. of Electrical Engineering\\
Texas A\&M University \\
\texttt{levimcclenny@tamu.edu} \\
\And
Mulugeta A. Haile \\
Army Research Lab \\
Aberdeen Proving Ground, MD \\
\And
Vahid Attari \\
Dept. of Materials Science \\
Texas A\&M University \\
\And
Brian M. Sadler\\
Army Research Lab \\
Adelphi, MD \\
\And
Ulisses M. Braga-Neto \\
Dept. of Electrical Engineering \\
Texas A\&M University \\
\And
Raymundo Arroyave \\
Dept. of Materials Science \\
Texas A\&M University \\
}

\begin{document}
\abovedisplayskip=0pt
\belowdisplayskip=0pt
\maketitle

\begin{abstract}

In many real-world applications of deep learning, estimation of a target may rely on various types of input data modes, such as audio-video, image-text, etc. This task can be further complicated by a lack of sufficient data. Here we propose a Deep Multimodal Transfer-Learned Regressor (DMTL-R) for multimodal learning of image and feature data in a deep regression architecture effective at predicting target parameters in data-poor domains. Our model is capable of fine-tuning a given set of pre-trained CNN weights on a small amount of training image data, while simultaneously conditioning on feature information from a complimentary data mode during network training, yielding more accurate single-target or multi-target regression than can be achieved using the images or the features alone. We present results using phase-field simulation microstructure images with an accompanying set of physical features, using pre-trained weights from various well-known CNN architectures, which demonstrate the efficacy of the proposed multimodal approach.

\end{abstract}

\section{Introduction}
\begin{figure}[hbt]
    \centering
    \centering
    \subfigure[CNN Estimation on an image-only domain. Here, $\textbf{F}:\textbf{X}\rightarrow \textbf{y}$ is a ConvNet used to predict a value in the target domain]{\includegraphics[width=.38\linewidth, height=1.5in]{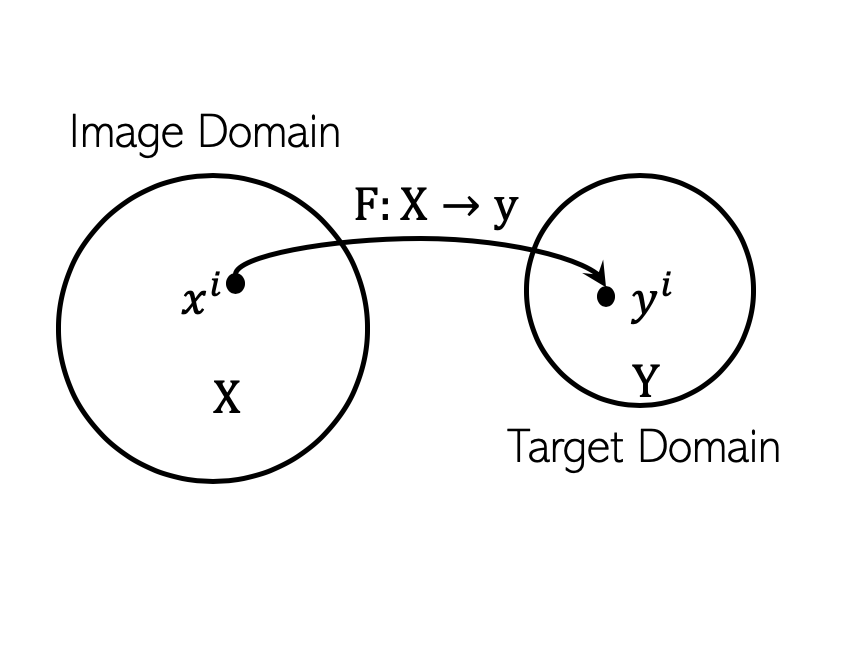}}
    \hspace{3em}
    \centering
    \subfigure[Estimation on a multimodal image-descriptor domain. Here $\textbf{F}:\textbf{X}_1, \textbf{X}_2\rightarrow \textbf{y}$ is the proposed DMTL-R estimator]{\includegraphics[width=.38\linewidth, height=2.0in]{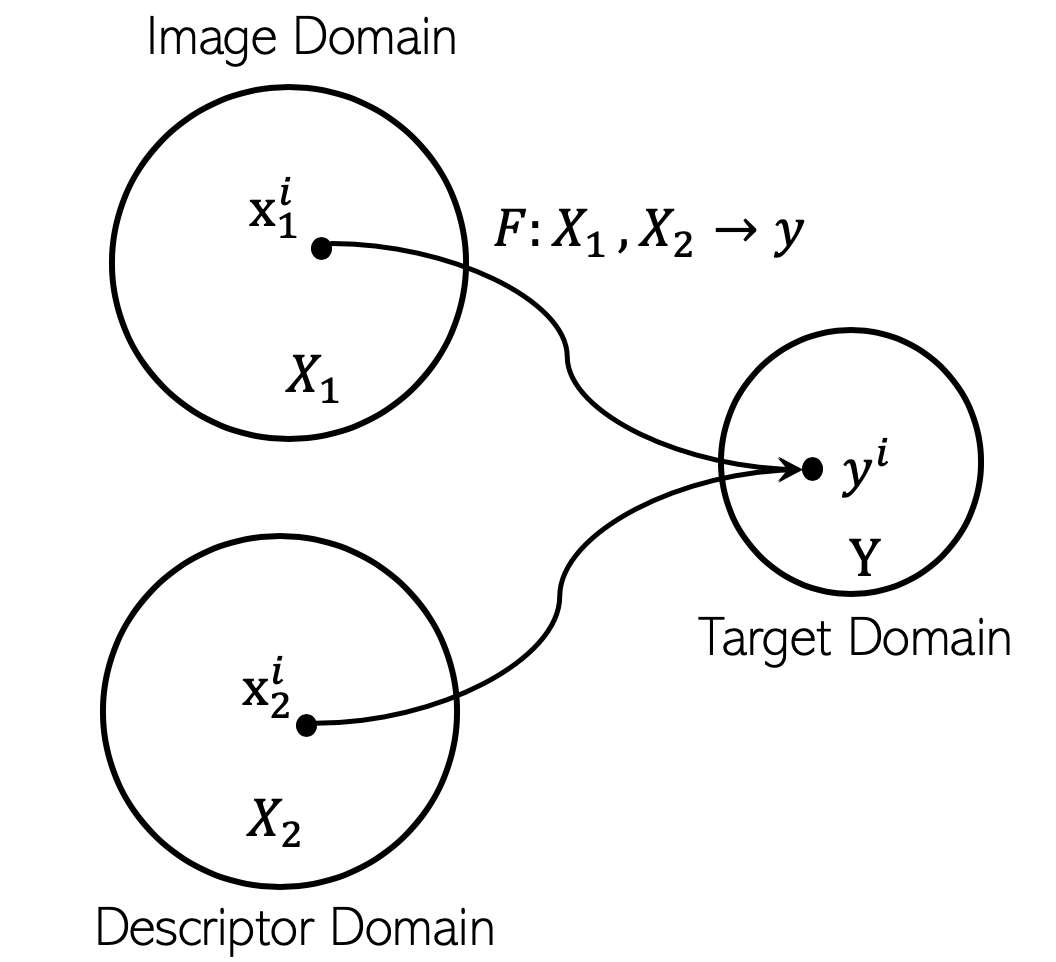}}
    \caption{Single-domain input vs. multi-domain input for multimodal regression. In (b) the images are combined with a corresponding vector describing the image in question in another data domain}
    \label{fig:domain}
\end{figure}

Consider the following problem - shown in Figure~\ref{fig:domain} - in which we desire to apply deep regression in an application domain in which a ConvNet has not been trained, and there exists additional data which is hypothesized to assist in this regression task. 
In this instance, assume further that there does not exist sufficient data to train a new ConvNet from random initializations. In this article, we provide a multimodal architecture which takes advantage of model fine-tuning and transfer learning to overcome a lack of sufficient training data. The result is a regression approach that combines images and descriptive statistics which can be effectively trained on a modestly sized dataset.

This paper makes the following major contributions: 
\begin{itemize}
    \item Develops a multimodal deep learning regression methodology that incorporates CNN-based image featurization conditioned on descriptive statistics of the input images
    \item Shows efficacy of the multimodal transfer-learned regression methodology in a data-poor application domain 
\end{itemize}

Our approach yields a transfer-learned regressor with better residual $R^2$ than image-only regression alone, and presents an algorithm that allows for the combination of image and descriptive statistics that can be applied to a large variety of scientific domains. 
This provides a potential approach to a significant question in deep learning - how to effectively incorporate interpretable a priori descriptive information about an image-based system into an estimation task - which we address in this work via multimodal deep learning.

We validate this approach on images of phase-field simulated microstructures with accompanying descriptive statistics about the corresponding material system, and present this concatenated information to an estimator that seeks to regress parameters about the image in question. Phase-field simulation microstructures are extensively used to study the physical and mechanical properties of materials and provide a relevant example in the context of this algorithm, as microstructure data is expensive to obtain.

\section{Related Work}

\paragraph{Multimodal Learning}

The concept of combining different domains of input into a single estimator for the desired learning task has been of distinct interest in the last decade~\cite{ramachandram2017deep, gao2020survey}. Applications include speech classification using audio and video input~\cite{ngiam2011multimodal}, tagging or labeling of images using features and textual information~\cite{srivastava2012multimodal}, and numerous others. 
This framework is useful in approaching the problem we outline in Figure~\ref{fig:domain}, as it allows for combinations of different types of input data, which together describe a similar location in the input space. 
It is worth clarifying that what could be considered \textit{descriptive information} about the image in~\cite{srivastava2012multimodal} are image captions, i.e. textual information. 
In this work, we refer to descriptive information as \textit{numerical} vectors describing the image.
We apply this framework for a regression task, specifically combining the image and descriptive input domains into a single estimator. 

\paragraph{Multi-Source Domain Adaptation}
Domain Adaptation is considered a branch in the broad area of transfer learning~\cite{pan2009survey, sun2015survey}. Specifically, multi-source domain adaptation addresses the question of multiple sensor inputs to an estimator in which there is only one target domain, and how to effectively leverage information from another domain in this new application. This can typically be accomplished either via adding up all the data sources into a single source or by training classifiers on each branch and aggregating those results for a final estimation. This particular approach, most generally, aggregates results from classifiers trained on the same \emph{type} of data, from various sources, not necessarily different domains (or modes), such as what we see in Figure~\ref{fig:domain}. In this work, we seek to develop a \emph{multimodal} domain adaptive regressor using transfer learned networks.

\paragraph{Multi-Input Transfer Learning}
As mentioned above, domain adaptation is a broad branch of transfer learning. Recently, multi-source domain adaptation using pre-trained networks has caught much attention, as it generates models that allow for a smaller amount of training data for the model in question~\cite{lee2019learning}. Recent work has been done in the multi-source transfer learned estimator space~\cite{li2019multi, xu2012multi, guo2016representation}, using text-text and text-image combinations to perform classification. Most of these works propose frameworks that are built in an ensemble fashion, i.e. trained independently, and then the best classifier selected. In this work, we address the task of regression specifically, using a single estimator (built around a transfer-learned network) with synchronous training of parameters.

\paragraph{Deep Regression}
Recently, image-based deep regression has become a rapidly advancing application of deep learning~\cite{lathuiliere2019comprehensive}. Taking a CNN and applying it to the task of regression is a problem that has large implications in many areas of science and engineering. While the work in~\cite{lathuiliere2019comprehensive} is extremely thorough in the task of fine-tuning a CNN model using pre-trained weights, it does not incorporate the additional statistics described in this article as a component of their model. 
It does offer important insight into the task of using CNNs for regression, insight which is used in the formulation of the models in this article. 
We offer an extension of the deep regression task in~\cite{lathuiliere2019comprehensive} by taking additional descriptive information about images and incorporating that information into the estimation.

The proposed Deep Multimodal Transfer-Learned Regression (DMTL-R) algorithm is novel in that it requires the training of only one estimator, whereas most work related to multimodal transfer learned estimation requires the training of multiple estimators. 
Additionally, the entire model is built of connected layers, allowing backpropagation to flow through the entire network at once, removing any sort of selection requirement seen in other multi-source transfer learning-based regressors. 
Moreover, the results presented in section~\ref{sec:results}, which are gathered from a data set with a modest number of samples, indicate that our approach to multimodal regression is accurate and efficient in a data-poor environment.

\section{Development of the Model}

\paragraph{Fine-Tuning the ConvNet}

We define a ConvNet as a nonlinear function approximator that maps an image input to a target label (for prediction) or target value 
(for regression). The ConvNet is trained on input-target realizations $x \in \textbf{X}$ and $y \in \textbf{Y}$. Here we can say that the image-target pairs exist in the ILSVRC-2012 dataset~\cite{imagenet_cvpr09}, more commonly known as ImageNet. We'll call this domain $G$. 
Once trained, a ConvNet makes a prediction $\overline{\textbf{y}}$ for an input image or batch of images $\textbf{x}$. We define a prediction on a ILSVRC-2012 trained ConvNet $\overline{\textbf{F}} : \textbf{X} \rightarrow y$ as

\begin{equation}
  \overline{y}(x) = \overline{F}(x, w_T), \ x,y \in G
\end{equation}
  
where $G$ is the ILSVRC-12 domain, $x$ and $y$ are the image and target domains respectively, and $w_T$ are the trained weights of the network. 
In most applications, these weights are trained specifically to best featurize the images in the training domain of the ConvNet. 
These weights are unique to the layer to which they belong, and we can index them as $w_i$, $i \in \{1,n\}$ where $n$ is the number of trainable layers in the network. 
Model fine-tuning typically takes advantage of the subset of layers used to featurize an image, i.e. excluding the fully-connected layers in the VGG16 architecture. 
Here we can subset those weights as the image featurization weights $w_f \in w_T$. 
We generate a new neural network around the $w_f$ layers, with trainable weights $w_{FT}$, and create a featurization $u$ which captures the features of a new domain $\textbf{H} \notin \textbf{G}$. We can define a new function 

\begin{equation}
  u(x_H) = \overline{F}^*(x_H, w_f, w_{FT})
\end{equation}

Here $\overline{F}^*$ is trained via backpropagation in a similar sense to the original model, but the featurization layers $w_f$ are held constant - this creates a mapping of the activations of the image to a new trainable set of fully-connected layers $w_{FT}$.

\paragraph{Conditioning on Descriptive Statistics} 

We define a descriptive variable of arbitrary length $n$ as some associated vector describing the image data input to the ConvNet.

\begin{figure}[hbt]
    \centering
    \includegraphics[width=.72\linewidth]{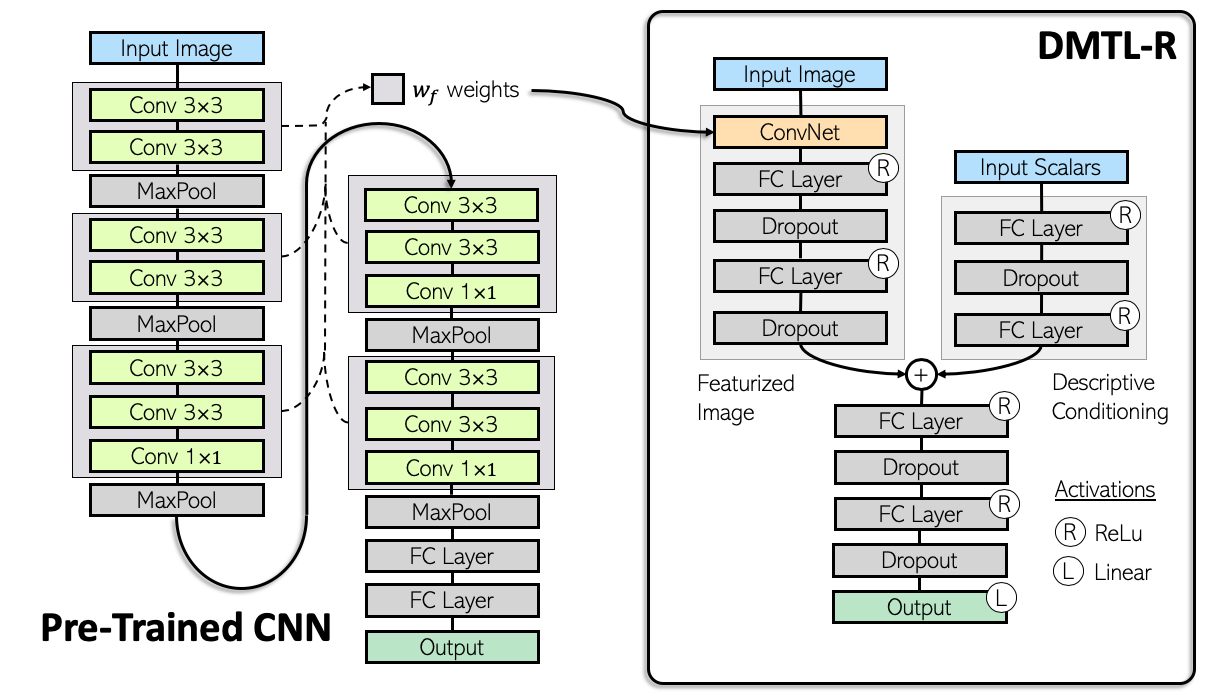}
    \caption{Proposed DMTL-R Estimator architecture. The left image is a block diagram of the VGG16 architecture, with its featurization weights $w_f$ highlighted. These are an example of featurization weights that can be used in the DMTL-R (right) to estimate the target value at the output layer.}
    \label{fig:block}
\end{figure}

In Figure~\ref{fig:domain}(b) we see the descriptor domain $\textbf{X}_2$ where a vector realization $x_2 \in \mathbb{R}^n$ is used as a descriptive statistic to condition the regression output of the multimodal regressor outlined in Figure~\ref{fig:block}.
We generate a small fully-connected network with trainable weights $w_{MLP}$ to allow the featurization of the conditioning statistics and allow for the optimal MSE regressor via backpropagation through the whole network. We will call this featurization component $v$, defined as

\begin{equation}
  v(x_2) = F(x_2, w_{MLP})
\end{equation}

Featurized image components and descriptive conditioning statistics are then concatenated in a fully-connected layer, which does not affect the differentiation of backpropagation. This combines the values and allows for addition of a final few fully-connected layers, with a dropout\cite{srivastava2014dropout} component added between most fully-connected layers induce regularization in the model via random masking of nodes. The weights $w_r$ of these last few layers are also trainable and result in the final regression. The output layer has a linear activation function, while all the intermediate layers use ReLu activation. 

We are left with a final regression algorithm $r(x_1, x_2)$ which accounts for image and descriptive statistics at sample point $i$ as follows:

\begin{align}
   r_i(x_1, x_2)  &= F (u(x_{1}^i), v(x_{2}^i), w_r) 
   \\ &= F(x_{1}^i, x_{2}^i, w_f , w_{FT}, w_{MLP}, w_r)
   \label{eq:model}
\end{align}

This is optimized via backpropagation, in the form of batch gradient descent against an MSE loss function 
\begin{align}
  \mathcal{L}_{batch} &= \frac{1}{n}\sum_i^{n}(y_T-\overline{y})^2 = \frac{1}{n}\sum_i^{n}(y^i_T-F (u(x_{1}^i), v(x_{2}^i), w_r))^2
  \\  &= \frac{1}{n}\sum_i^{n}(y^i_T-F(x_{1}^i, x_{2}^i, w_f , w_{FT}, w_{MLP}, w_r)
  )^2
  \label{eq:loss}
\end{align}

where $n$ is the batch size, $i \in \{1,n\}$ is a sample of that batch, and $\overline{y}$ and $y_T$ are the predicted and actual target values for that sample, respectively. All the trainable weights $w_{FT},w_{MLP}$ and $w_r$ are updated through the entire combined regressor using the final target prediction value $r_i(x_1, x_2)$. The intermediate functions $u(x_1), v(x_2)$ are not individually optimized.

\section{Training the Model}
\subsection{Training Data}
The training images used in this study are of material microstructures generated from a sweep of input physical coefficients to physics-based  phase-field simulations, which are a powerful tool in materials science to predict complex evolution kinetics in materials processes~\cite{steinbach2013phase}. 
The images utilized for this example are visual depictions of material phase separation during processing (also known as spinodal decomposition) and are results from the phase-field simulations. 
Our \textit{descriptive statistics} in this example are 18-tuple sets of continuous input processing parameters used for the phase-field simulation. These tuples act as physical parameters to a nonlinear set of coupled partial differential equations which are computationally difficult to analyze and must be solved numerically via Fourier spectral method (or another similar numerical solver) across many CPU cores - potentially hundreds.
Combining the two sets of inputs, we have an input data set composed of image data and corresponding vectors of numerical values. 
The target (output) values to be regressed are 6-tuple physical characteristic outputs from the phase field simulation, such as min/max compositions and chemical potential. These are simulated outputs from the phase-field solution that we seek to regress from the image data and corresponding input vectors using the above outlined DMTL-R approach. The results are discussed for various regression trials in section~\ref{sec:results}.

\begin{figure}[hbt!]
\centering
    \includegraphics[width=.7\linewidth]{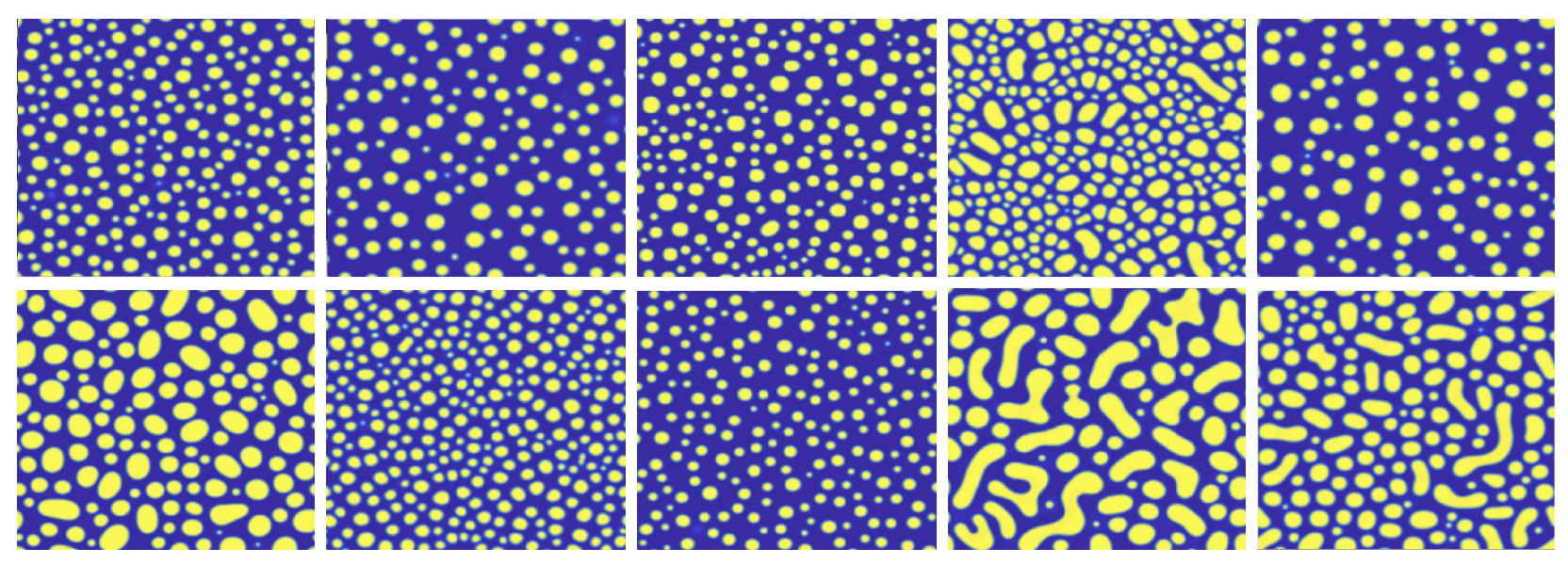}
  
\caption{Microstructure images generated with phase-field simulation and input to the combined image-parameter regressor. The images are featurized using ConvNet architectures and conditioned with descriptive statistics. The original image sizes are 512x512x3.}
\label{fig:data}
\end{figure}

The training/test data consists of 2500 images of fully spinoidally decomposed microstructures obtained from the open phase-field microstructure database~\cite{attari2020uncertainty, kunselman2020semi}.\footnote{http://microstructures.net/} Spinoidally decomposed images were chosen as they are the most visually diverse and informative images. 2500 images is a very small dataset for a CNN architecture, and the fully-connected architecture described in this review has a few million trainable parameters, i.e. $p >> n$. It is worth noting, however, that this is far fewer trainable parameters than most commonly used CNN architectures~\cite{chollet2015keras}. We account for the $p>>n$ phenomenon in the model by introducing dropout to prevent overfitting via sparsity induction. This is in addition to utilizing the transfer-learned ConvNet architecture, which on its own is a method of to proceed when using large deep learning methodologies with insufficient data to train a full-scale CNN model from random initializations. 

The dataset is split randomly into train and test sets, with 2/3 of the data for training and 1/3 for testing, or \char`\~1675/825. This split is done randomly and independently for each trial.

\begin{figure}[hbt]
    \centering
    \includegraphics[width=.8\linewidth]{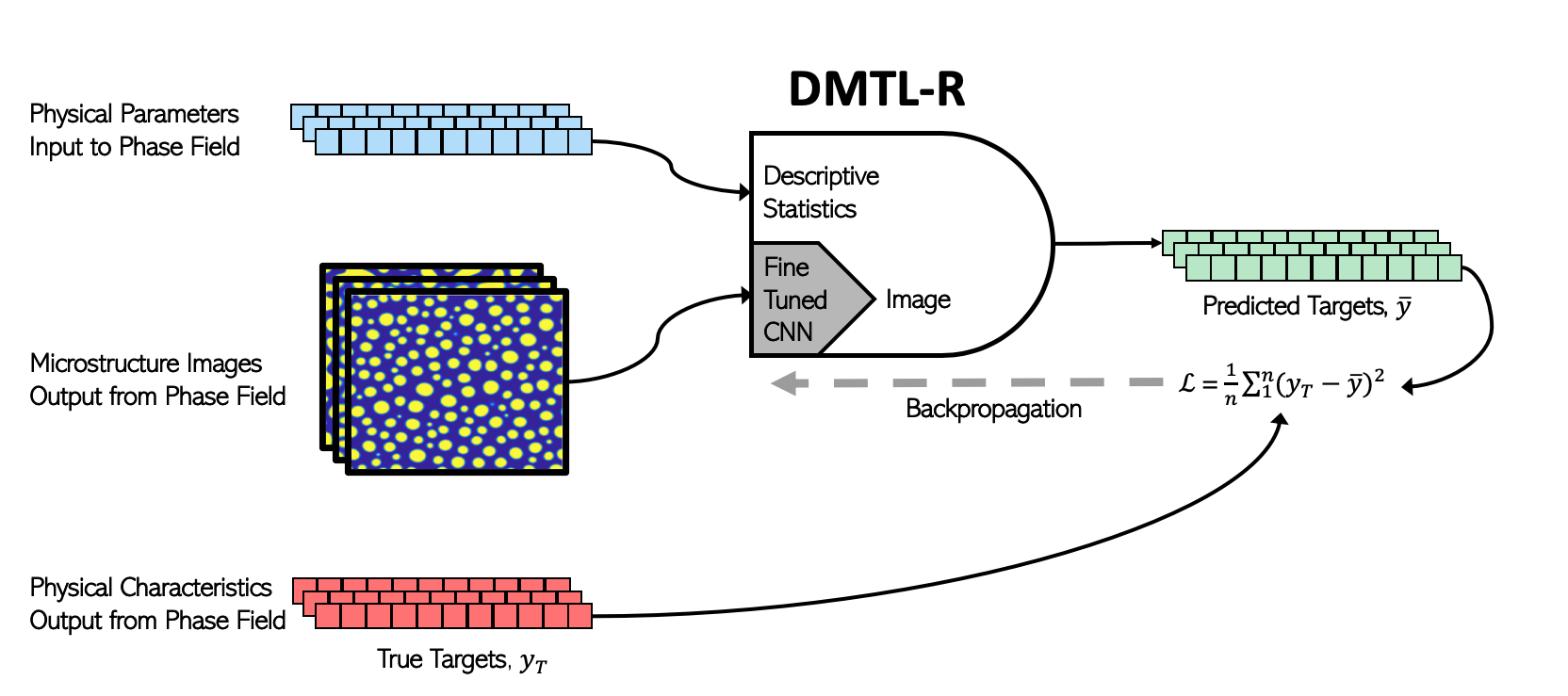}
    \caption{Training flow of the proposed Deep Multimodal Transfer-Learned Regressor (DMTL-R), the internals of which are shown in Figure~\ref{fig:block}}
    \label{fig:train}
\end{figure}

\subsection{Experimental Setup}
Once the available data is split into train and test sets, the images are resized (compressed) to the required input size (224x224 or 299x299, depending on the architecture) and mean pixel values are removed from each channel, as is common practice to centralize the image pixel distributions for ImageNet-based tasks and provide added stability in training. The descriptive statistics and the output target scalar values are mean-corrected and scaled to unit variance. Training iterations are reported in the next section for 20 training epochs. Adam optimization~\cite{kingma2014adam} was utilized for parameter optimization during training, with learning rate ranging from $.0001 < lr < .001$ with moderate decay. Batch size was set to 32 for all training comparisons in table~\ref{tab:res}. The model was built in Keras~\cite{chollet2018deep, rosebrock_CNN, rosebrock_Multi} and trained using a single Nvidia V100 GPU made available through an Amazon Web Service (AWS) P3 instance. With this setup, it takes 3-4s per epoch to train the DMTL-R network. A full implementation of this architecture has been made publicly available by the authors.\footnote{https://github.com/levimcclenny/multimodal\_transfer\_learned\_regression}

\section{Results} \label{sec:results}
\subsection{Model Fine-Tuning for CNN-Based Regression}\label{sec:regression}\label{sec:CNN}

The VGG16 architecture shown on the left side of Figure~\ref{fig:block} was used as a feature extractor for the microstructure images shown in Figure~\ref{fig:data} and trained as a regressor with 3 additional fully-connected layers of size $[1000, 100, n_{output}]$. The final output layer is given a linear activation function, while the intermediate layers use ReLu to induce nonlinearity to the estimation.

\begin{figure}[hbt]
    \centering
    \includegraphics[width=.9\linewidth]{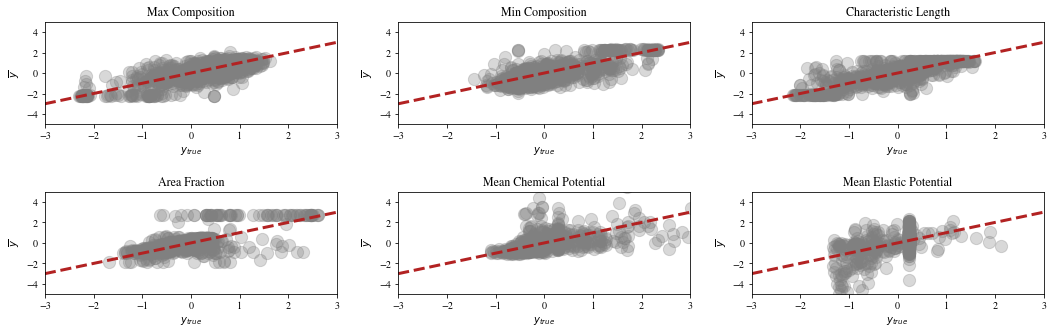}
    \caption{True vs. predicted estimates from CNN-based feature extraction using VGG16-based model fine-tuning. Max Composition, Min Composition, etc. are physical output characteristics of the spinoidal decomposition simulations.}
    \label{fig:CNNresults}
\end{figure}

While this architecture does accomplish the goal of predicting a target from a transfer learned ConvNet featurization, it can be determined from inspection of Figure~\ref{fig:CNNresults} that the images on their own provide only weak training for the regression task. This does not mean that the model is ineffective. Rather, it implies that there is not enough information in the images alone to effectively train a regressor, which on its own is somewhat useful information about the physical problem at hand. To aid in this problem we condition on descriptive features, as shown in Figure~\ref{fig:block}, which is a focal point of this work. This generates a multimodal estimator that includes multi-domain input information for a single point in the input space. The results shown in Figure~\ref{fig:CNNresults} should be used as the baseline for comparison for the results in sections to follow, and are quantitatively tabulated for comparison in Table~\ref{tab:res}. 

\subsection{Single-Target DMTL Regression}
\begin{figure}[hbt]
    \centering
    \includegraphics[width=.9\linewidth]{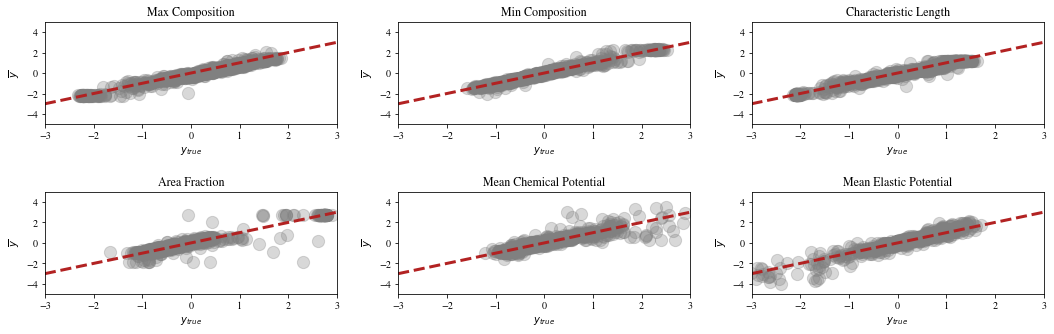}
    \caption{True vs. predicted estimates from single-target DMTL regression}
    \label{fig:results}
\end{figure}

\begin{figure}[hbt!]
    \centering
    \includegraphics[width=.9\linewidth]{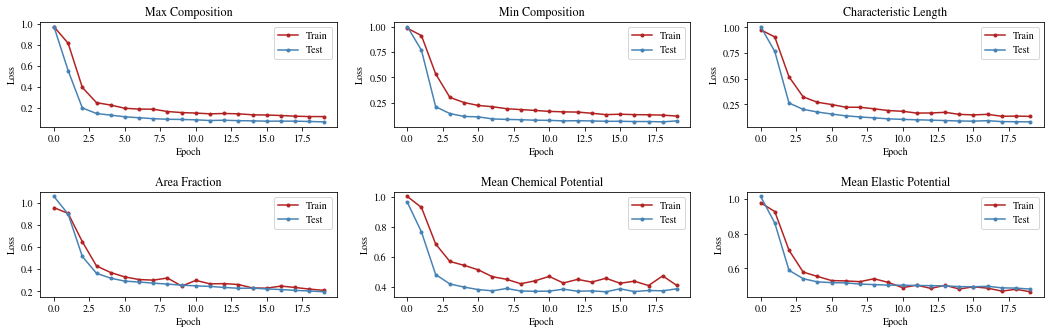}
    \caption{Single-target DMTL-R regression train and test set loss over 20 training epochs}
    \label{fig:loss}
\end{figure}

The first validation experiment conducted with DMTL-R is that of single-target regression. The CNN-based featurization from section~\ref{sec:regression} is paired with descriptive statistics as suggested in the architecture in Figure~\ref{fig:train}.
The final layer in the model from Eq.~\ref{eq:model}, with a linear activation function and MSE loss from Eq.~\ref{eq:loss}, was given a node size of $1$ and trained to regress a single parameter at a time. 
The DMTL-R results are shown in Figure~\ref{fig:results}, with plots of the MSE loss through training epochs shown in Figure~\ref{fig:loss}. 

We note here that the predicted estimate vs. true target values are, on average, very accurate with a reasonably low residual MSE. 
It's important to note that the regression targets, as well as the input descriptive vectors, were scaled using a standard scaler, i.e. subtracting the mean and scaled to unit variance.
Therefore, the MSE values are somewhat arbitrary when discussing the predicted values of the parameters themselves. 
The MSE metric is, however, useful in comparing methodological and architectural differences, as well as providing a very stable loss function for training the DMTL-R network. 

Figure~\ref{fig:loss} shows the training and test loss for each training epoch in the single-target regression case. We see that the model architecture, despite being in a position where overfitting could be preeminent in the model, does a reasonably good job in maintaining generalization to the test set. After 20 training epochs, the loss values approximately converge and the test set error does not exceed the training set error, which is typically interpreted as an indication of overfitting in the model. This would suggest that, despite the temptation of overfitting when training in a $p>>n$ environment, ConvNet transfer learning paired with standard sparsity induction techniques (such as dropout) do a reasonably good job of maintaining generalization while creating a strong multimodal regressor. 

\subsection{Multi-Target DMTL Regression}
\begin{figure}[hbt]
    \centering
    \includegraphics[width=.96\linewidth]{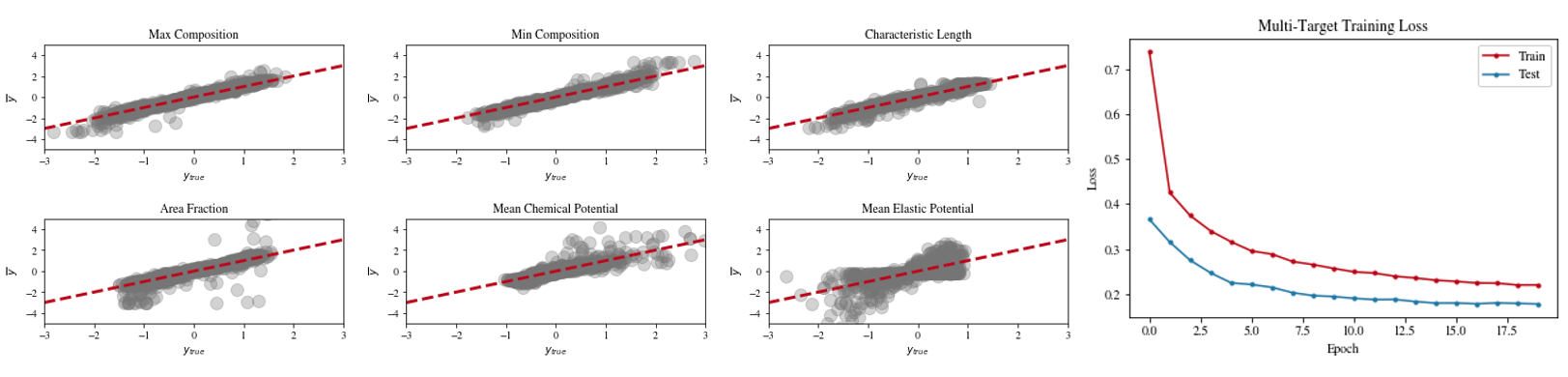}
    \caption{Estimates and train/test loss over 20 training epochs for the multi-target regressor}
    \label{fig:multi}
\end{figure}

Multi-target regression has been a longstanding topic in traditional pattern recognition spaces. One of the biggest strengths of deep learning is that increasing target diminsionality is as straightforward as increasing the number of nodes in the output layer. Here we extend the DMTL-R regressor to multi-target regression, in this case regressing all 6 parameters shown in Figures~\ref{fig:results} at the same time. This results in little additional computational cost and can potentially have a positive effect on the accuracy of the individual dimensions estimated by the regressor, as seen in Table~\ref{tab:res}. The loss and regression results of the multi-target regression are shown in Figure~\ref{fig:multi}.

Table~\ref{tab:res} outlines interpretation metrics for the regressor models, including the $R^2$ values for the respective regressors. This $R^2$ is from a linear fit for each true parameter vs. it's estimates, on the test set, for each output variable regressed. The slope, in all instances, is very close to $1$, as anticipated. The $R^2$ value is listed as a metric by which we can assess the goodness-of-fit of the regressor. These values are listed in table~\ref{tab:res} with 95\% confidence intervals derived from multiple independent trials with independent splits of train/test sets for each trial, to validate generalization.
We can see here that the multimodal DMTL-R regressor fares very well in regressing the target parameters, with most $R^2$ values well over 0.90. This method of analysis provides an intuitive illustration of a large statement - that the multimodal DMTL-R regressor provides a good, sufficiently general estimate to test data. Further, analysis of the trend of losses in Figures~\ref{fig:loss} and~\ref{fig:multi} also suggests that the models are able to maintain generality across test sets. 

\begin{table}[ht!]
\resizebox{\textwidth}{!}{
\begin{tabular}{c||c|c|c||c|c|c||}

\multicolumn{1}{l||}{}   & \multicolumn{3}{c||}{\textbf{Single-Target Regression}} & \multicolumn{3}{c||}{\textbf{Multi-Target Regression}} \T\B\\
\cline{2-7}   & \textbf{DMTL-R}  & Image Only   & Statistics Only   & \textbf{DMTL-R} & Image Only     & Statistics Only  \T\B  \\
\cline{2-7} Target Index     & \multicolumn{6}{c||}{\textbf{ResNet50}} \T\B  \\ 
\hline
1 & 0.979 $\pm$ .0040 & 0.375 $\pm$ .1194  &  0.969 $\pm$ .0011  & \textbf{0.981} $\pm$ .0020 & 0.660 $\pm$ .0630 & 0.963 $\pm$ .0060 \T\\ 
2  & 0.984 $\pm$ .0058  & 0.290 $\pm$ .1136  & 0.974 $\pm$.0009  & \textbf{0.985} $\pm$ .0019  & 0.688 $\pm$ .0485 & 0.966 $\pm$ .0063  \\ 
3  & 0.978 $\pm$ .0065  & 0.567 $\pm$ .1157 & 0.963 $\pm$ .0012  & \textbf{0.979} $\pm$ .0020 & 0.608 $\pm$ .0621 & 0.956 $\pm$ .0038  \\
4   & 0.899 $\pm$ .0246  & 0.507 $\pm$ .0673  & 0.895 $\pm$ .0053  & \textbf{0.925} $\pm$ .0059  & 0.657 $\pm$ .0435 & 0.886 $\pm$ .0152  \\
5   & 0.780 $\pm$ .0420  & 0.316 $\pm$ .0208  & 0.763 $\pm$ .0019  & \textbf{0.823} $\pm$ .0108  & 0.405 $\pm$ .0408 & 0.732 $\pm$ .0186  \\
6   & \textbf{0.943} $\pm$ .0084  & 0.505 $\pm$ .0405  & 0.751 $\pm$ .0084  & 0.736 $\pm$ .0183  & 0.522 $\pm$ .0575 & 0.705 $\pm$ .0037 \B \\
\hline\hline
\multicolumn{1}{c||}{}    & \multicolumn{6}{c||}{\textbf{VGG16}} \T\B \\
\hline
1   & 0.973 $\pm$ .0039  & 0.797 $\pm$ .0210 & 0.969 $\pm$ .0011   & \textbf{0.980} $\pm$ .0022   & 0.808 $\pm$ .0254 & 0.963 $\pm$ .0060  \T \\
2 & 0.976 $\pm$ .0026  & 0.805 $\pm$ .0171   & 0.974 $\pm$.0009  & \textbf{0.983} $\pm$ .0021    & 0.817 $\pm$ .0212  & 0.966 $\pm$ .0063\\
3  & 0.965 $\pm$ .0028  & 0.821 $\pm$ .0270    & 0.963 $\pm$ .0012   & \textbf{0.976} $\pm$ .0029  & 0.828 $\pm$ .0289 & 0.956 $\pm$ .0038   \\
4   & 0.926 $\pm$ .0080 & 0.684 $\pm$ .0509   & 0.895 $\pm$ .0053   & \textbf{0.939} $\pm$ .0059 & 0.735 $\pm$ .0390 & 0.886 $\pm$ .0152 \\
5  & \textbf{0.843} $\pm$ .0342 & 0.340 $\pm$ .1703     & 0.763 $\pm$ .0019  & 0.803 $\pm$ .0366   & 0.512 $\pm$ .1046 & 0.732 $\pm$ .0186  \\
6   & \textbf{0.786} $\pm$ .0196   & 0.620 $\pm$ .0312  & 0.751 $\pm$ .0084  & 0.752 $\pm$ .0231  & 0.786 $\pm$ .0196      & 0.705 $\pm$ .0037 \B  \\
\hline\hline
                        & \multicolumn{6}{c||}{\textbf{InceptionV3}}  \T\B  \\
\hline
1  & \textbf{0.983} $\pm$ .0035 & 0.541 $\pm$ .0607 & 0.969 $\pm$ .0011  & 0.957 $\pm$ .0043 & 0.439 $\pm$ .0668 & 0.963 $\pm$ .0060 \T \\
2  & \textbf{0.987} $\pm$ .0018 & 0.546 $\pm$ .0694 & 0.974 $\pm$ .0009  & 0.956 $\pm$ .0049 & 0.439 $\pm$ .0695 & 0.966 $\pm$ .0063 \\
3  & \textbf{0.978} $\pm$ .0020 & 0.509 $\pm$ .0379  & 0.963 $\pm$ .0012  & 0.950 $\pm$ .0048 & 0.446 $\pm$ .0521 & 0.956 $\pm$ .0038 \\
4  & \textbf{0.903} $\pm$ .0077 & 0.530 $\pm$ .0313 & 0.895 $\pm$ .0053  & 0.853 $\pm$ .0066 & 0.488 $\pm$ .0469 & 0.886 $\pm$ .0152 \\
5  & \textbf{0.794} $\pm$ .0174 & 0.315 $\pm$ .0285 & 0.763 $\pm$ .0019  & 0.728 $\pm$ .0334 & 0.321 $\pm$ .0423 & 0.732 $\pm$ .0186 \\
6  & \textbf{0.943} $\pm$ .0053 & 0.425 $\pm$ .0343 & 0.751 $\pm$ .0084  & 0.777$\pm$ .0305 & 0.347 $\pm$ .0261 & 0.705 $\pm$ .0037 \B                             
\end{tabular}
}
\caption{Comparison of $R^2$ values (with 95\% confidence intervals) for single-target and multi-target regression via DMTL-R (our approach) described in Figure~\ref{fig:train}. $R^2$ values are after 20 training epochs. Results are derived from each of the ResNet~\cite{he2016deep}, VGG16~\cite{simonyan2014very}, and Inception~\cite{szegedy2016rethinking} architectures. }
\label{tab:res}
\end{table}

For comparison, in Table~\ref{tab:res}, \textbf{Image-only} statistics are fine-tuned CNN regressors trained without descriptive statistics with varying transfer learned architectures as shown, which are described briefly in section~\ref{sec:CNN}. \textbf{Statistics-only} columns refer to a regressor trained without the image component. These are fully-connected network regressors with hidden layer sizes $ [ d_{input}, 100, 100, 50, d_{output} ] $, which mimics the input to the descriptive statistic component of the DMTL-R model. This is trained with identical hyperparameters for learning rate, epochs, train/test split ratios, and training batch size. These results are shown in each row only for comparison and do not vary with CNN architecture. 

It is interesting to note that, in most instances, the DMTL-R approach was able to improve upon the image or statistic only fits. 
This confirms the hypothesis that an estimator which is capable of including both modes of information, such as the DMTL-R model, is a more exhaustive and complete model. 
With how DMTL-R is trained and the ability to generalize, we believe DMTL-R is the superior model for the task of multimodal image-descriptor regression in the presence of small training data. We demonstrate with an improvement in $R^2$ by a range of 4-5\% over statistics-only and 59-117\% over image-only regression.

\section{Conclusion}
In this paper, we presented a Deep Multimodal Transfer-Learned Regressor (DMTL-R) for predicting target parameters in data-poor domains.   
The inputs to the regressor are images and a corresponding $n$-tuple of statistics containing information we know to be true about the image from another data domain. The suggested DMTL-R approach, built around a pre-trained CNN, featurizes the image then conditions its features with corresponding descriptive statistics. We studied a materials science application, regressing 6 dimensions of output target parameters from input images and 18-tuple input statistics.
We found that with the available small training sample, our approach results in better regression accuracy ($R^2$) of target parameters than a similar model trained over images or descriptive parameters alone. Here we have demonstrated the efficacy of DMTL-R on materials data, but the model could be extended to other data-poor domains such as healthcare, climatology, and beyond.

\begin{ack}
The authors would like to acknowledge the support of the D$^3$EM program funded through NSF Award DGE-1545403. The authors would further like to thank the US Army CCDC Army Research Lab for their generous support and affiliation. 
\end{ack}

\bibliographystyle{unsrt}  
\bibliography{neurips_2020}


\end{document}